\pdfoutput=1

\documentclass[11pt]{article}
\usepackage[table]{xcolor}
\usepackage[multiple]{footmisc}
\usepackage{multirow}
\usepackage{tabularx}
\usepackage{mdframed}
\usepackage{caption}
\usepackage{fontawesome}

\usepackage[final]{acl}

\usepackage{times}
\usepackage{latexsym}
\usepackage{longtable, booktabs, multirow}
\usepackage{tablefootnote}

\usepackage[T1]{fontenc}

\usepackage[utf8]{inputenc}

\newcolumntype{C}[1]{>{\centering\arraybackslash}p{#1}}
\newcommand{\namedref}[2]{\hyperref[#2]{#1~\ref*{#2}}}
\newcommand{\sectionref}[1]{\namedref{Section}{#1}}
\newcommand{\tableref}[1]{\namedref{Table}{#1}}
\newcommand{\figureref}[1]{\namedref{Figure}{#1}}

\usepackage{microtype}

\usepackage{inconsolata}

\usepackage{graphicx}
\usepackage{amsmath}
\usepackage{listings}
\usepackage{float}

\newcommand{\sys}{\textsc{System}}
\newcommand{\resp}{\textsc{Response}}
\newcommand{\evid}{\textsc{Evidence}}
\newcommand{\facts}{\textsc{Facts}}
\newcolumntype{Y}{>{\raggedright\arraybackslash}X}

\definecolor{sepcol}{gray}{0.85}
\definecolor{exrule}{gray}{0.80}           

\newcolumntype{L}{>{\raggedright\arraybackslash}X}
\newcolumntype{R}{>{\raggedright\arraybackslash}p{0.34\linewidth}}
\newcommand{\exampledivider}{\arrayrulecolor{exrule}\midrule[0.6pt]\arrayrulecolor{black}}

\definecolor{promptgray}{gray}{0.96}
\newcommand{\promptcell}[1]{\cellcolor{promptgray}{\ttfamily\footnotesize #1}}

\usepackage{stackengine}
\definecolor{boltgold}{RGB}{255,160,0}
\definecolor{boltedge}{gray}{0.15}
\definecolor{vfsblue}{RGB}{10,55,110} 
\definecolor{vsgray}{gray}{0.25} 
\usepackage{contour}
\contourlength{0.18ex}
\newcommand{\fastVerify}{%
  \stackinset{c}{0pt}{c}{0pt}
  {\scalebox{0.6}{\textcolor{boltgold}{\contour{white}\faBolt}}}{\textcolor{vfsblue}{\Large\faSearch}}}

\newcommand{\name}{\textcolor{vfsblue}{\textsc{VeriFastScore}}}
\newcommand{\veriscore}{\textcolor{vsgray}{\textsc{VeriScore}}}
\newcommand{\factscore}{\textcolor{vsgray}{\textsc{FactScore}}}
\newcommand{\safe}{\textcolor{vsgray}{\textsc{Safe}}}
\newcommand{\namer}{\textcolor{vfsblue}{\textsc{VeriFastScorer}}}
\newcommand{\veriscoredata}{\textcolor{vsgray}{\textit{VeriScore data}}}
\newcommand{\tulupersonas}{\textcolor{vsgray}{\textit{Tulu3 Personas}}}

\title{ \fastVerify \hspace{0.1cm} \name: Speeding up long-form factuality evaluation}

\author{
  Rishanth Rajendhran$^{1}$ \quad
  Amir Zadeh$^{2}$ \quad
  Matthew Sarte$^{2}$ \quad
  Chuan Li$^{2}$ \quad
  Mohit Iyyer$^{1}$ \\
  $^{1}$University of Maryland, College Park \quad
  $^{2}$Lambda Labs \\
  $^{1}$\texttt{\{rishanth, miyyer\}@umd.edu} \quad $^{2}$\texttt{\{amirali.zadeh, matt.sarte, c\}@lambdal.com}
}

\begin{document}
\maketitle
\begin{abstract}

Metrics like \factscore\ and \veriscore\ that evaluate long-form factuality operate by decomposing an input response into atomic claims and then individually verifying each claim. While effective and interpretable, these methods incur numerous LLM calls and can take upwards of 100 seconds to evaluate a single response, limiting their practicality in large-scale evaluation and training scenarios.
To address this, we propose \name, which leverages synthetic data 
to fine-tune Llama3.1 8B  for \emph{simultaneously} extracting and verifying all verifiable claims within a given text based on evidence from Google Search. We show that this task cannot be solved via few-shot prompting with closed LLMs due to its complexity: the model receives \textbf{$\sim$4K} tokens of evidence on average and needs to concurrently decompose claims, judge their verifiability, and  verify them against noisy evidence. However, our fine-tuned \name\ model demonstrates strong correlation with the original \veriscore\ pipeline at both the  example level ($r=\mathbf{0.80}$) and  system level ($r=\mathbf{0.94}$) while achieving an overall speedup of \textbf{6.6$\times$} (9.9$\times$ excluding evidence retrieval) over \veriscore. To facilitate future factuality research, we publicly release our \name\ model and synthetic datasets.\footnote{Code and data available at \url{https://github.com/RishanthRajendhran/VeriFastScore}.}
\end{abstract}
\section{Introduction}
\label{sec:intro}

Modern methods to evaluate the factuality of long-form generation, such as \factscore~\cite{min2023factscorefinegrainedatomicevaluation}, \safe~\cite{wei2024longformfactualitylargelanguage}, and \veriscore~\cite{veriscore}, rely on a pipelined approach that uses LLMs to decompose an input text into ``atomic'' claims and then verify each claim against retrieved evidence documents. While this pipeline results in interpretable outputs that correlate strongly with human annotations of factuality, it is also \emph{slow} ~\cite{liu-etal-2025-towards}, requiring many calls to large  models to evaluate a text, which limits its usability in both evaluation and training (e.g., as a reward model in RLHF).

In this paper, we propose a significantly faster alternative: a single-pass factuality evaluator that simultaneously performs claim decomposition and verification over a long-form input. Our method, \name, is trained to jointly extract all verifiable claims from a response and assess their factuality against retrieved evidence. To train \name, we collate sentence-level evidence with claim-level outputs produced by \veriscore, a high-quality general-domain factuality evaluation metric, into a synthetic dataset used to fine-tune Llama3.1 8B Instruct~\cite{grattafiori2024llama3herdmodels}.

\begin{figure*}[ht!]
    \centering
    \includegraphics[width=\linewidth]{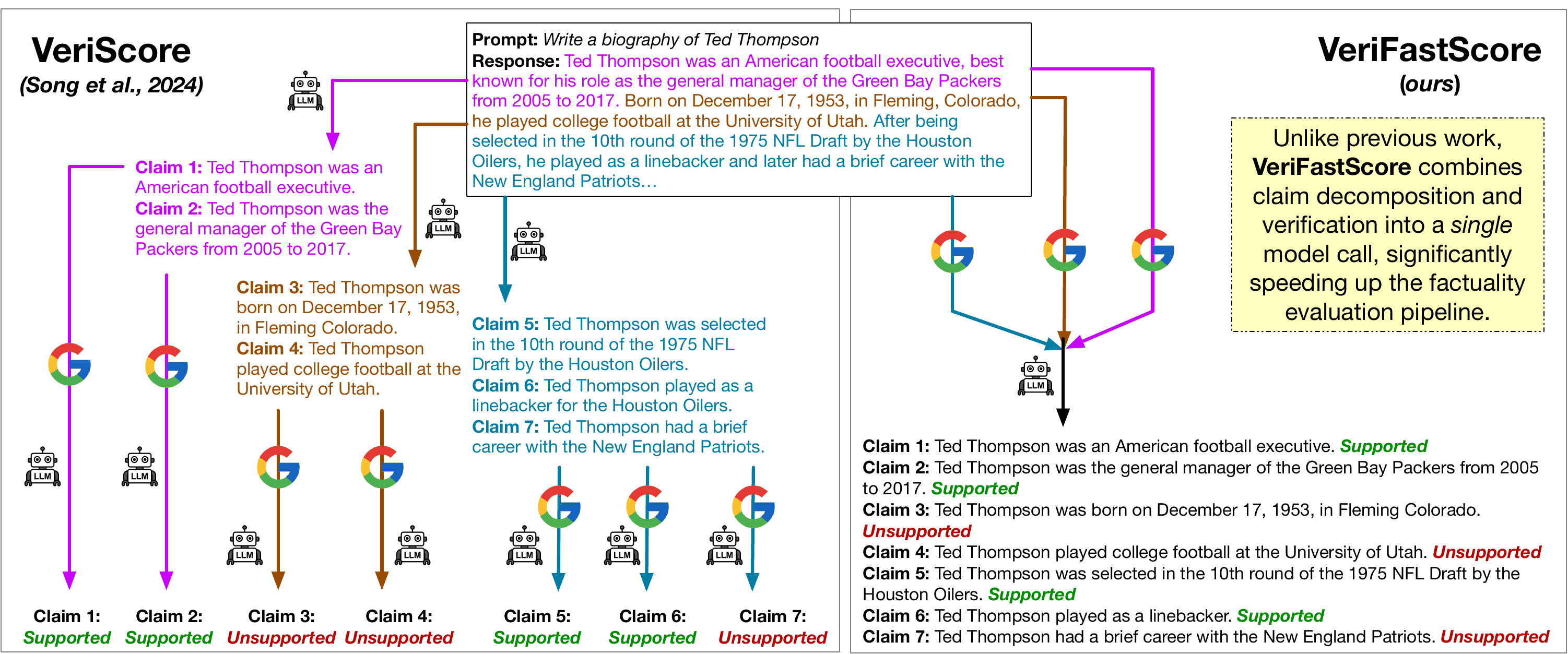}
    \caption{\emph{(left)} The \veriscore\ pipeline \cite{veriscore} involves a separate call to a trained model (or LLM API) to decompose every sentence of the long-form response. More model calls are issued for each decomposed claim to verify them against claim-level evidence retrieved from Google Search. In contrast, our \name\ method retrieves \emph{sentence-level} evidence from Google Search, combines it together with the response into a single long-context prompt, and outputs a list of claims and labels with just a \emph{single call} to a trained model.}
    \label{fig:sec1_veriscore_vs_fastfact_pipeline}
    \vspace{-7pt}
\end{figure*}

Each training example contains a response to be evaluated as well as a corresponding evidence context that consists of web search results retrieved by querying individual sentences in the response. \name\ is then asked to output a list of verifiable claims from the response, each labeled as \texttt{Supported} or \texttt{Unsupported} based on the evidence (see Figure~\ref{fig:sec1_veriscore_vs_fastfact_pipeline}). This task is non-trivial: the model must \emph{internally} decontextualize the response, identify atomic claims, resolve coreferences, and assess support against an evidence context spanning up to \textbf{4K} tokens of potentially noisy information. Standard few-shot prompting of powerful closed models like GPT-4o struggles with this complexity, achieving a Pearson correlation of only 0.33 with \veriscore\ despite high API costs.

Despite these challenges, \name\ matches the quality of \veriscore\ closely while offering substantial improvements in both cost and runtime efficiency: it achieves a Pearson correlation of $\mathbf{0.80}$ (Pearson r, p<0.001) with \veriscore's scores, a \textbf{9.9$\times$} modeling speedup and a \textbf{6.64$\times$} overall speedup in wall-clock time, reducing both modeling and retrieval latency while maintaining the metric's interpretability. Human evaluation corroborates the high quality of outputs from \name. We release \name\ and our synthetic training dataset to support efficient, interpretable factuality evaluation and facilitate its use in alignment settings.



\section{Building a faster factuality metric}



In this section, we first describe the limitations of existing factuality metrics like \veriscore, whose separate decomposition, retrieval, and verification stages require several calls to LLMs to complete for a single response. These limitations motivate our design and development of \name, which combines all steps in the existing factuality evaluation pipeline into a single model call. To enable such a fast evaluation, we generate synthetic data from \veriscore\ and use it to fine-tune a language model on \emph{response-level} decomposition and verification.

\begin{table*}[t]
\fontsize{8}{10}\selectfont
\centering
\resizebox{\textwidth}{!}{%
  \begin{tabular}{@{}p{2cm}c c c c c c c c@{}}
    \toprule
    \textbf{Prompt Source}
      & \textbf{Split}
        & \textbf{\# Responses}
          & \textbf{Avg.\ Claims/Response}
            & \multicolumn{3}{c}{\textbf{Avg.\ Token Length}}
              & \multicolumn{2}{c}{\textbf{Claim Verification Label (\%)}} \\
    \cmidrule(lr){5-7} \cmidrule(lr){8-9}
    & & & 
      & \textbf{Response}
        & \textbf{Evidence}
          & \textbf{Facts}
            & \textbf{Supported}
              & \textbf{Unsupported} \\
    \midrule

    \multirow{2}{*}{\veriscoredata}
      & Train & 4082 & 16 & 298 & 4381 & 271 & 68 & 32 \\
      & Test  & 1815 & 16 & 301 & 4466 & 278 & 68 & 32 \\

    \midrule

    \multirow{2}{*}{\tulupersonas}
      & Train & 4860 & 20 & 366 & 3755 & 334 & 59 & 41 \\
      & Test  & 4083 & 20 & 370 & 3714 & 329 & 59 & 41 \\

    \bottomrule
  \end{tabular}%
}
\caption{\textbf{Overview of the datasets used.} For each data source and split (train/test), we report the total number of model responses, the average number of verifiable claims per response, average token lengths \tablefootnote{Token lengths are estimated using Llama 3.1 Instruct tokenizer.} of model responses, consolidated evidence, and consolidated list of verifiable claims extracted by \veriscore\ from model responses, and the percentage of claims labeled by \veriscore\ as supported and unsupported by retrieved evidence. Labels are more balanced in the dataset collected during this work using prompts from \tulupersonas\ dataset.}
\label{tab:support_stats}
\end{table*}

\subsection{The ``decompose and verify'' pipeline for factuality evaluation}
Recently-proposed metrics for long-form factuality evaluation such as \factscore\ and \safe\ are implemented by the following multi-stage pipeline:
\begin{enumerate}
    \item \textbf{Claim extraction:} decompose the long-form model response into ``atomic'' (i.e., short) claims
    \item \textbf{Evidence retrieval:} collect evidence for every claim individually by retrieving relevant documents from a datastore or search engine
    \item \textbf{Claim verification:} mark each atomic claim as either \emph{supported} or \emph{unsupported} by the evidence collected for that claim
\end{enumerate}

In \factscore\ and \veriscore, the first and third stages are implemented by either open-weight or closed LLMs (e.g., GPT-4), while SAFE's decomposition and verification stages are implemented by prompting closed models from OpenAI or Claude.

\paragraph{\veriscore, our starting point:}
Unlike \factscore~\cite{min2023factscorefinegrainedatomicevaluation} and \safe~\cite{wei2024longformfactualitylargelanguage}, which attempt to verify \emph{all} atomic claims in a model's response using external evidence, the decomposition stage of \veriscore\ focuses solely on \emph{verifiable} claims—defined as 
\begin{quote} 
\noindent \underline{Verifiable claims} describe a single \textit{event} or \textit{state}\footnote{Event: change of state, for example, ``Jensen Huang founded NVIDIA in 1993 in California, U.S.'' State: for example, ``Westborough is a town in Worcester, MA.''} with all necessary modifiers (e.g., spatial, temporal, or relative clauses) that help denote entities or events in the real world. They should be verifiable against reliable external sources (e.g., Wikipedia). They should exclude personal experiences, subjective opinions, hypotheticals, suggestions, advice, or instructions.
\citep{Maienborn, EandS} \end{quote}
Like \safe, \veriscore\ also relies on Google Search to retrieve evidence in the second stage of the pipeline. We use \veriscore\ as a starting point due to its increased domain generality compared to prior long-form factuality metrics.

\subsection{Issues with VeriScore} 

\veriscore\ evaluates factuality using a multi-stage pipeline that proceeds sentence-by-sentence through a model response. For each sentence, it extracts verifiable claims using a claim extraction model, where the full response is included in the input to ensure proper decontextualization. Each extracted claim is then used as a query to retrieve evidence via Google Search. Finally, a claim verification model checks the factuality of each claim against the retrieved evidence.

While this careful decomposition and step-wise verification ensures precision, it incurs significant computational cost. On average, a single response of 14 sentences leads to 23 extracted claims (see \tableref{tab:cost_estimate}). Each response thus results in:
\begin{itemize}
    \item 14 calls to the claim extraction model (one per sentence),
    \item 23 Google searches (one per claim),
    \item 23 calls to the claim verification model.
\end{itemize}
This totals roughly 60 model or API calls per response. Processing a single response takes approximately 100 seconds, making the method impractical for real-time or large-scale evaluation \cite{liu-etal-2025-towards, veriscore}.

\subsection{\name's Integrated Approach}

Given advances in instruction following and long-context reasoning in modern LLMs, we investigate whether it is possible to perform claim decomposition and verification in tandem using a single model pass. A central challenge in such an approach is the order of operations: traditional pipelines perform decomposition before retrieval in order to resolve coreferences, identify verifiability, and construct precise search queries. By bypassing decomposition, we must ensure that evidence retrieval remains effective despite potential ambiguity or context dependence in the original response.

\paragraph{Evidence Retrieval Without Claim Decomposition:} 
To retrieve evidence without prior knowledge of claims, we treat each sentence in the model response as a search query. We use the SERPER API\footnote{\url{serper.dev}}, as in \veriscore, to obtain the top 10 search results per sentence. Snippets from all queries are concatenated to form a consolidated evidence context. This method leverages redundancy in web search and the semantic breadth of full-sentence queries to compensate for the lack of explicitly formulated claims.

\paragraph{Simultaneous Claim Decomposition and Verification:}
With the consolidated evidence and the full model response as inputs, \name\ performs both claim extraction and verification in a single forward pass. This removes the need for intermediate steps and per-claim processing, offering a significantly more efficient pipeline.

Despite the lack of decomposition at the retrieval stage, we find that \name\ maintains strong performance. On average, each response contains $\sim$23 verifiable claims and is associated with $\sim$186 evidence snippets, totaling over 4k tokens. These are processed together, with \name\ identifying and verifying each claim directly against the evidence context. As shown in \tableref{tab:dataset_performance} and \tableref{tab:execution_time}, \name\ achieves high factual precision and recall with respect to \veriscore, while reducing inference time per response by over 6$\times$.

\paragraph{Score calculation:} To score a model’s response, we adapt the metrics used in \veriscore. Factual precision of a model response $r$ is defined as:\\
\[
P(r) = \frac{S(r)}{|C|}
\]
where $C$ is the set of all verifiable claims in $r$ and $S(r)$ denotes the number of verifiable claims in $r$ that are supported by the consolidated evidence $E$. \\
\paragraph{} Furthermore, to prevent models from gaming the evaluation, it is important to also consider the number of verifiable claims produced by the model in its response. Thus, we also compute factual recall as follows:\\
\[
R(r) = min(1, \frac{|C|}{K}) 
\]
where $K$ is the median number of verifiable claims across model responses. \\

We compute $F_1@K(r)$:

\begin{equation*}
  F_1@K(r) =
      \begin{cases}
      \frac{2P(r)R_K(r)}{P(r) + R_K(r)} & \text{if } S(r) > 0\\
      0 & \text{if } S(r) = 0
    \end{cases}
\end{equation*}

The final score for a model $\mathcal{M}$ across a dataset $X$ is:
\begin{equation} 
\label{eq:factuality_score}
\text{\name} = \frac{1}{|X|} \sum_{x\in X} F_1@K(\mathcal{M}_x)
\end{equation}

\begin{table}[t]
\centering
\small
\begin{tabularx}{\linewidth}{@{}lY@{}}
\toprule
\textbf{Section} & \textbf{Content (see \autoref{tab:prompt-format} for full text}) \\ \midrule
\sys & \promptcell{You are trying to verify factuality by extracting atomic, verifiable claims\ldots
Output: \textless fact\textsubscript{i}\textgreater:\; \{Supported, Unsupported\}.
If none, output ``No verifiable claim.''} \\[3pt]

\resp & \promptcell{\ldots In the summer of 1215, a pivotal moment in the course of Western political and legal history unfolded in England.\ldots} \\[3pt]

\evid & \promptcell{The Magna Carta is the charter of English liberties granted by King John on June 15, 1215,\ldots} \\[3pt]

\facts & \promptcell{1. The Magna Carta is considered a pivotal moment in Western political and legal history: \textbf{Supported}}\par
2. \dots\\
\bottomrule
\end{tabularx}
\caption{Heavily-truncated example of the \textbf{prompt format} used to fine-tune \name\ for response-level factuality evaluation. The model receives a \textsc{system} prompt, \textsc{response}, and \textsc{evidence}, and it is asked to generate a list of \textsc{facts} that consists of 
 verifiable claims and their labels.}
\label{tab:prompt-format}
\end{table}

\subsection{Generating Synthetic Training Data for \name}



\paragraph{Data Sources and Motivation:} 
To train \name, we began with data collected for human evaluation during the development of \veriscore\ (see \veriscoredata\ in \tableref{tab:support_stats}) consisting of prompts sourced from LongFact \cite{wei2024longformfactualitylargelanguage}, AskHistorian, ELI5 \cite{xu2023criticalevaluationevaluationslongform}, ShareGPT \cite{vicuna2023}, FreshQA \cite{vu2023freshllmsrefreshinglargelanguage} etc. Early experiments with subsets of this data of different sizes ranging between 0.5K and 5K revealed a clear trend of improved model performance with increasing training data size. Motivated by this, we aimed to scale up training via a synthetic dataset generated using the \veriscore\ pipeline. Specifically, we targeted prompts likely to elicit factual claims, enabling us to generate high-quality supervision for claim decomposition and verification.

\paragraph{Prompt Selection and Dataset Choice:}
We selected prompts from the \tulupersonas\ dataset \cite{lambert2025tulu3pushingfrontiers}, which contains diverse and high-quality instructions across domains intended to test model's instruction-following ability and adaptation to user intent. This dataset was chosen because it aligns well with our goal of generating model responses that contain factual, verifiable content. We used GPT-4o-mini to identify a subset of prompts likely to elicit factual claims\footnote{More details in \ref{appendix:tulu3-sieve}}. A manual inspection of 200 filtered prompts showed that over 90\% did indeed lead to factually dense responses.

\begin{table*}[t]
\centering
\fontsize{14}{18}\selectfont
\resizebox{\textwidth}{!}{%
\begin{tabular}{@{}p{4.5cm}p{4.5cm}C{1.75cm}C{1.5cm}ccccp{1cm}@{}}
\toprule
    \textbf{Evidence Granularity} 
  & \textbf{Model} 
  & \textbf{Precision $\uparrow$} 
  & \textbf{Recall $\uparrow$} 
  & \multicolumn{3}{c}{\textbf{Claim Accuracy (\%)}} 
  & \textbf{Correlation b/w factuality scores $\uparrow$} \\ 
\cmidrule(lr){5-7}
& & & & \textbf{Correct $\uparrow$} & \textbf{Incorrect $\downarrow$} & \textbf{Missing $\downarrow$} & \\
\midrule

\multirow{2}{*}{claim‐level}
  & GPT-4o few-shot ICL     & 0.30    & 0.31    & 18.1    & \textbf{7.2}    & 73.2    & 0.28    \\
& \name  & \textbf{0.87} & \textbf{0.90} & \textbf{71.6} & 15.5 & \textbf{12.3} & \textbf{0.87} \\
\midrule

\multirow{2}{*}{sentence-level}
& GPT-4o few-shot ICL     & 0.38    & 0.34    & 19.3    & \textbf{8.2}    & 71.2    & 0.33    \\
& \name  & \textbf{0.83} & \textbf{0.86} & \textbf{66} & 17.6 & \textbf{15} & \textbf{0.80} \\
\bottomrule
\end{tabular}%
}
\caption{\textbf{Performance metrics across test splits:} \name\ produces claims and  verification labels that are largely consistent with \veriscore, achieving a strong corelation of \textbf{0.80} even with sentence-level evidence, while  significantly outperforming the  GPT-4o few-shot prompting baseline. \textit{Evidence Granularity} denotes whether evidence was retrieved using full sentences or claims: claim-level includes \veriscoredata\ and \tulupersonas\ data while sentence-level includes \tulupersonas\ data only. \textit{Precision} and \textit{Recall} measure overlap between claims identified by \name\ and \veriscore. Under \textit{Claim Accuracy}, \textit{Correct}, \textit{Incorrect (Label)}, and \textit{Missing (Claim)} capture agreement, label mismatches, and omissions relative to \veriscore. A small fraction of instances were marked \textit{Erroneous} (not shown) due to ambiguities in automatic evaluation (see \ref{appendix:auto-eval-appendix}). \textit{Correlation} reports Pearson $r$ between factuality scores (see  Eq. \ref{eq:factuality_score}) assigned by \name\ and \veriscore. For all Pearson's correlation scores reported in this paper, p<0.001, unless stated otherwise.}
\label{tab:dataset_performance}
\end{table*}

\paragraph{Data Collection Pipeline:}
From the filtered prompts, we generated responses using a selection of LLMs: GPT-4o \cite{openai2024gpt4ocard}, Llama3.1 8B, Gemma2 7B \cite{gemmateam2024gemma2improvingopen}, Mistral-v2 7B \cite{jiang2023mistral7b}, and Qwen2 7B \cite{yang2024qwen2technicalreport}. We collected a total of about 9K prompt-response pairs, ensuring diversity across model architectures and outputs. Each response was processed through the \veriscore\ pipeline to decompose it into verifiable claims and assign verification labels based on evidence retrieved using those claims.

\paragraph{Discrepancy between evidence used during training vs. test-time:}
A key difference between \veriscore\ and \name\ is the granularity at which evidence for claim verification is retrieved. As illustrated in \figureref{fig:sec1_veriscore_vs_fastfact_pipeline}, \veriscore\ first extracts verifiable claims from the model response and then uses them as search queries to retrieve evidence for claim verification. This is in contrast to the setting in \name\ where evidence for claim verification is collected even before claim decomposition. This is achieved by using the sentences in the model response to collect search results which are then consolidated together for verification. 

To train and evaluate \name\ model, we opted to keep the claim-level evidence collected by \veriscore\ during the synthetic data generation pipeline outlined earlier for two reasons: 
\begin{itemize}
    \item it allowed us to reuse the unmodified and validated \veriscore\ pipeline to ensure high-quality data, and
    \item it promotes robustness by training the model to verify claims against arbitrary but relevant evidence sets.
\end{itemize}
Additionally, we also collected sentence-level evidence for the Tulu3-personas subset of our training and test data, in-line with the true setting of \name. We evaluate the model's ability to generalize to test-time evidence conditions in Section~\sectionref{sec:results}.

\paragraph{Evidence Overlap and Label Robustness:}
To assess whether our training setup, consisting largely of claim-level evidence, is compatible with test-time inference, we analyzed the overlap between evidence gathered using these two different approaches. We found that 17\% of claim-level retrieved URLs are also present in the sentence-level evidence, and about 60\% of claims have at least one matching URL in the sentence-level evidence.\footnote{Note that we only count exact URL matches; semantic overlap is likely higher.} 

\paragraph{Empirical 
proof of soundness:} Despite this modest overlap, we observed a strong correlation between verification labels and model-level factuality scores produced by \name\ and \veriscore\ (discussed in \sectionref{sec:results}), suggesting that the model generalizes well across evidence types. This empirical agreement indicates that the discrepancy in evidence provenance has limited practical impact, and supports the validity of using claim-level supervision for training.

\subsection{Evaluation Metrics}

Since we motivate \name\ as a speedy replacement to the multi-stage pipeline of \veriscore, we want its outputs to be closely aligned with outputs from \veriscore. We evaluate \name\ using three main metrics:
\begin{itemize}
        \item \textbf{Claim Precision:} Computed as the proportion of claims produced by \name\ that are also produced by \veriscore,
    \item \textbf{Claim Recall:} Computed as the proportion of claims produced by \veriscore\ that are also produced by \name,
    \item \textbf{Claim Accuracy:}
        \begin{itemize}
            \item \textbf{Correct:} Percentage of claims produced by both \name\ and \veriscore\ with the same verification label,
             \item \textbf{Incorrect Label:} Percentage of claims produced by both \name\ and \veriscore\ with the differing verification label,
             \item \textbf{Missing Claim:} Percentage of claims produced by \veriscore\ but not by \name\ (analogous to Claim recall), 
        \end{itemize}
    \item \textbf{Pearson's Correlation:} Correlation coefficient between factuality scores of model responses produced by \name\ and by \veriscore.
\end{itemize}

\paragraph{Automatic Evaluation:} 
During early experiments, we found that when using exact match metrics, comparing model-extracted claims and verification labels directly to reference outputs, even minor surface-level differences (e.g., paraphrasing, slight reordering, or omission of non-essential modifiers) led to false negatives, thereby underestimating the model's actual performance (see Appendix \tableref{tab:dataset_performance}). For example, \name\ achieved an exact match claim accuracy of 23.7\% on the test set, versus 71.6\% with paraphrase-aware evaluation.\footnote{We re-report \tableref{tab:dataset_performance} with exact match metrics in Appendix.}

While exact match metrics were retained during training for their efficiency and stability, we adopted a more nuanced automatic evaluation for the test set. Specifically, we used GPT-4o-mini as a judge to compare the model outputs against gold reference annotations and assess semantic overlap. This helped capture more accurate performance trends. The evaluation setup and prompts are detailed in Appendix~\ref{appendix:auto-eval-appendix}.

\subsection{Training the \name\ Model}
\label{sec:training}
We train \name\ using a two-stage finetuning procedure on synthetic data generated via the \veriscore\ pipeline. The base model is Llama3.1 8B Instruct, trained first on claim-level evidence and then on a mixture of claim- and sentence-level evidence to better match the test-time setting. This mixed-evidence setup improves robustness and mitigates the observed performance drop when relying solely on sentence-level evidence (see \tableref{appendix:dataset_performance}). 

All training examples consist of a model-generated response and retrieved evidence, with supervision derived from \veriscore’s annotated claim lists and verification labels. Details of the multi-stage training setup are provided in Appendix~\ref{appendix:training}.

\section{Results}

\label{sec:results}
In this section, we describe the test set performance and computational cost of the trained \name\ model and compare it with \veriscore. 

\begin{figure}[htbp]
    \centering
    \includegraphics[width=\columnwidth]{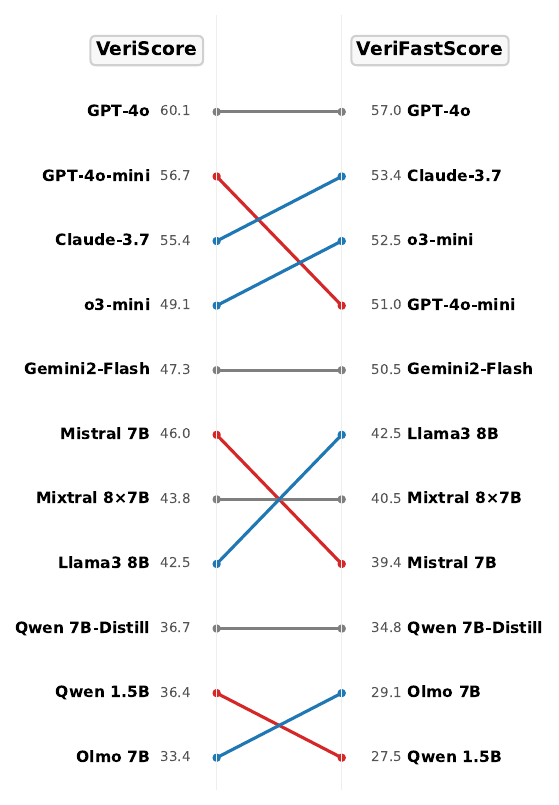}
    \caption{Model rankings produced by \veriscore\ and \name\ using a random subset of 100 prompts from the test split of Tulu3-Personas data. The rankings produced by both models are fairly consistent with each other despite some differences in order. There is a clear separation between open-weight and closed-weight models.}
    \label{fig:rankings}
\end{figure}

{\begin{table*}[t]
\small
\centering
\begin{tabularx}{\textwidth}{@{} L !{\color{sepcol}\vrule width 1pt} R @{}}
\toprule
\textbf{Example (truncated)} & \textbf{Remarks} \\ \midrule

\textbf{Response} … The “chimney’’ and “riverbank’’ serve as powerful metaphors — the chimney suggesting an unreachable goal while the riverbank signifies his current position …\par
\textbf{\name}: {\footnotesize(Unsupported) The main character in “Araby’’ imagines climbing a chimney to reach his desire.}\par
\textbf{\name}: {\footnotesize(Unsupported) The “riverbank’’ in “Araby’’ signifies the main character’s current position.}\par
\textbf{\veriscore}: {\footnotesize(Unsupported) The chimney symbolizes an unreachable goal.}\par
\textbf{\veriscore}: {\footnotesize(Unsupported) The riverbank symbolizes the protagonist’s current position.} &
This example illustrates the benefit of response-level extraction in \name\ over the sentence-focused \veriscore; \name\ better de-contextualises each claim. \\[4pt]
\exampledivider

\textbf{Response} … Navigating a rainforest can pose several challenges … unpredictable weather which can lead to sudden rainstorms or flooding …\par
\textbf{\name}: {\footnotesize(Supported) Unpredictable weather in a rainforest can lead to sudden rainstorms.}\par
\textbf{\veriscore}: {\footnotesize(Unsupported) Navigating a rainforest can lead to sudden rainstorms.} &
In this example, \name\ exhibits precise semantic alignment between extracted claim and model response while \veriscore\ does not.\\[4pt]
\exampledivider

\textbf{Response} … JSON-formatted letter quoting Admiral Chester W. Nimitz: “Courage is not the absence of fear …’’ …\par
\textbf{\name}: {\footnotesize(Unsupported) Admiral Chester W. Nimitz said, “Courage is not the absence of fear, but rather the judgment that something else is more important than fear.’’}\par
\textbf{\veriscore}: {\footnotesize No verifiable claim.} &
\name\ remains robust even when facts are buried in noisy, non-standard contexts (here, a JSON letter). \\[4pt]
\exampledivider

\textbf{Response} … The fast-paced editing, over-the-top physical comedy, and clever sound effects have become hallmarks of animation, influencing everything from anime to Pixar films …\par
\textbf{\name}: {\footnotesize(Unsupported) Over-the-top physical comedy is a hallmark of animation.}\par
\textbf{\veriscore}: {\footnotesize(Supported) Classic cartoons influenced the use of over-the-top physical comedy in animation.} &
\textbf{Spurious inference via clause-conflation:} \name\ sometimes glues fragments into an unsupported cause–effect claim. \\

\bottomrule
\end{tabularx}
\caption{\textbf{Qualitative analysis:} \name\ exhibits generalization and robustness at test time due to diverse training, but still inherits some undesirable behavior from \veriscore.}
\label{tab:qualitative-ex-main}
\end{table*}
}

\subsection{Main results}

\paragraph{High correlation between outputs from \name\ and those from \veriscore:} In \tableref{tab:dataset_performance}, we observe that \name\ outputs have both a high factual precision and factual recall with outputs from \veriscore\ as ground truth irrespective of the provenance of the evidence. There is a drop in claim accuracy when using sentence-level evidence which manifests itself as an increase in incorrect verification label and in missing claims. The higher percentage of missing claims can possibly be mitigated by more training on sentence-level evidence.  

\paragraph{GPT-4o Baseline:} For a baseline, we prompt GPT-4o with three few-shot demonstrations. Despite the high API cost (roughly \textbf{\$300} for our evaluation), GPT-4o is about \textbf{50\% less accurate} than \name\ which can even be run locally. A closer inspection revealed that GPT-4o failed to extract any verifiable claim from about 29\% of the instances. These results underscore the complexities involved in simultaneous claim extraction and verification. 

\subsection{Speedup}
\begin{table}[ht]
\centering
\small
\begin{tabular}{@{}
>{\raggedright\arraybackslash}p{3cm}   
>{\centering\arraybackslash}p{2cm}     
>{\centering\arraybackslash}p{2cm}        
@{}}
\toprule
& \multicolumn{2}{c}{\textbf{Average time spent (s)}} \\
\cmidrule(lr){2-3}
& \textbf{Evidence retrieval} & \textbf{Modeling}\\
\midrule
\veriscore
& 21.4 & 83.0 \\
\name
& 13.5 &  9.5 \\
\midrule
Speedup
& 1.9x     & 9.9x \\
\bottomrule
\end{tabular}
\caption{\textbf{Execution time:} \name\ spends significantly lesser time than \veriscore\ in the modeling stages i.e. extraction and verification stages. Overall, \name\ is 6.6x faster than \veriscore\tablefootnote{\label{estimate_mark}Estimated from 200 instances}.}
\label{tab:execution_time}
\end{table}

We measured the execution time for both the \name\ and \veriscore\ pipelines. By retrieving evidence at a coarser granularity, \name\ achieves a 39\% reduction in API costs and cuts retrieval time by almost half. Additionally, \veriscore\ takes significantly longer due to its sequential claim extraction and verification processes. \name\ is approximately 10 times faster in extracting and verifying claims without sacrificing performance (see Table \ref{tab:dataset_performance}). Overall, \name\ achieves a 6.64x speedup compared to \veriscore. Longer model responses further increase speedup, as \veriscore's sentence-by-sentence processing is particularly slow and expensive, in-part due to the larger number of SERPER queries \veriscore{} makes compared to \name{} (see Table \ref{tab:cost_estimate}).

\begin{table}[ht]
\centering
\small
\begin{tabular}{@{}
>{\raggedright\arraybackslash}p{2cm}   
>{\centering\arraybackslash}p{1.5cm}     
>{\centering\arraybackslash}p{1.5cm}     
>{\centering\arraybackslash}p{1.5cm}     
@{}}
\toprule
& \textbf{Retrieval time (\%)}
& \textbf{Average SERPER queries} 
& \textbf{Cost estimate\tablefootnote{Estimated using the tier priced at \$0.75/1000 queries} (\$)}\\
\midrule
\veriscore
& 20.5 & 23 & \$0.01725\\
\name
& 58.8 & 14 & \$0.0105\\
\bottomrule
\end{tabular}
\caption{\textbf{Cost estimates:} \name\ makes lesser search queries than \veriscore\ thereby reducing cost\textsuperscript{\getrefnumber{estimate_mark}}.}
\label{tab:cost_estimate}
\end{table}

\subsection{Ranking factuality of models with \name}

We randomly sample 100 prompts from the Tulu3-personas test set to obtain responses from 12 different LLMs:
\begin{itemize}
    \item \textbf{Closed-weight:}  GPT-4o, GPT-4o-mini, o3-mini \cite{o3mini}, Claude-3.7-Sonnet \cite{Claude3S}, Gemini-2.0-Flash \cite{gemini2025}
    \item \textbf{Open-weight:} Llama3.1 8B, Mistral v0.3 7B, Mixtral v0.1 8x7B, Qwen2.5 1.5B \cite{qwen2025qwen25technicalreport}, Deepseek-R1-distill Qwen 7B \cite{deepseekai2025deepseekr1incentivizingreasoningcapability}, Olmo2 7B \cite{olmo20252olmo2furious}
\end{itemize}

As shown in Figure~\ref{fig:rankings}, factuality scores produced by \name\ correlate strongly with those from \veriscore\ (Pearson $r = 0.94$, p=$1.1e^{-5}$), indicating alignment despite differing approaches. Given that \name\ operates over a single model pass and avoids per-claim retrieval, it delivers these rankings with significantly reduced latency and cost (see \tableref{tab:execution_time}, \ref{tab:cost_estimate}).

\section{Runtime Analysis}
\label{sec:parallel}

To more precisely quantify the efficiency difference, we implemented a parallelized
version of \veriscore{} and compared it directly with \name{}.
We distributed sentence-level claim decomposition and claim-level verification
across \textbf{8 GPUs} on a H100 node.

\paragraph{Parallelized \veriscore{} Implementation:} 
For a 100–instance subset of the test set, this setup reduced the
\emph{claim-decomposition} time to an average of \textbf{59.7s}
(compared to \textbf{253s} for the standard single-GPU implementation),
with a total modeling time (decomposition + verification) of
approximately \textbf{235s}.

\paragraph{Comparison with Unparallelized \name{}:} By contrast, \name{} processes the same 100 instances in only
\textbf{29s on a \emph{single} GPU} on the same H100 node, while performing both claim
extraction and verification in one pass.

\paragraph{Limits of Parallelization:} Although parallelization yields moderate speedups,
\veriscore{} remains fundamentally constrained by its
\emph{sequential pipeline}:
(1) claims must first be extracted sentence-by-sentence,
and (2) each extracted claim must then be verified individually.
These dependencies create repeated tokenization and input-formatting costs,
along with inter-process communication overhead,
all of which limit scalability and make efficient batching difficult.
In contrast, \name{} integrates decomposition and verification into a
single forward pass over the entire response and consolidated evidence,
eliminating intermediate stages and achieving consistent,
end-to-end efficiency gains.




\section{Qualitative Analysis}
\label{appendix:human-eval}
To better understand \name’s behavior, the first author conducted a small-scale human evaluation of 401 (response, evidence, claim, label) tuples, drawn from 21 test instances in the \tulupersonas\ dataset. The evaluation focused on three core aspects of \name’s output: claim extraction, claim verifiability, and verification accuracy.

\paragraph{Claim grounding and hallucination:}  
All 401 extracted claims were grounded in the original model response, with no instances of hallucinated or fabricated claims. This is a crucial property for a factuality metric, particularly in downstream use cases like hallucination detection and RLHF training where spurious claim hallucination could mislead supervision.

\paragraph{Extraction coverage and omissions:}  
\name\ was generally effective at identifying claims, especially in responses rich with factual content. However, it showed reduced reliability in responses that interweave factual and non-factual content, or contain few factual claims overall. Claim omission was observed in about 6 out of the 21 evaluated responses, though 4 of these cases involved fewer than four missing claims. Notably, in 3 of the omission cases, \name\ extracted claims only from the final few sentences of the response, missing earlier factual content—suggesting a positional bias or early truncation.

\paragraph{Unverifiable and ill-formed claims:}  
About 42 out of the 401 extracted claims were judged to be unverifiable due to missing key details (e.g., named entities, dates) in the original model response. In rare cases, \name\ also extracted claims from inherently unverifiable sentences, such as advice or vague generalizations. Ill-formed claims, those missing critical contextual elements, were rare (around 4 claims out of the total 401 claims), but they did alter the meaning of the original sentence in notable ways.

\paragraph{Verification accuracy:}  
\name\ was largely accurate in labeling claims against the provided evidence, with an estimated verification error rate of 91.5\%. Among the 34 incorrect verification labels, 25 of them were false positives: instances where \name\ incorrectly marked a claim as supported. These errors often arose when partial matches or topical overlaps misled the model, particularly in cases where evidence spanned multiple unrelated contexts or referenced similar entities in different time periods or locations.

\paragraph{Evidence quality and noise:}  
Finally, we observed that the retrieved evidence was often noisy, especially in cases where the original response contained few or vague factual claims. This is likely due to the sentence-level retrieval strategy used by \name, which may issue underspecified queries and produce loosely related results. These limitations in evidence quality can contribute both to omission in extraction and errors in verification.

\section{Related work}
\label{sec:related_work}

\paragraph{\textbf{Factuality evaluation via decomposition and verification}}
\textsc{FactScore} computes factual precision over atomic claims \citep{min2023factscorefinegrainedatomicevaluation}, while \textsc{SAFE} uses GPT-4 for claim-wise verification \citep{wei2024longformfactualitylargelanguage}. \veriscore\ extracts only verifiable claims and uses Google Search to verify them \citep{veriscore}. Other systems like \textsc{FacTool} \citep{chern2023factoolfactualitydetectiongenerative}, \textsc{RARR} \citep{gao2023rarrresearchingrevisinglanguage}, and \textsc{CoVe} \citep{elazar2021measuringimprovingconsistencypretrained} operate similarly albeit with document-level or retrieval-augmented reasoning.

\paragraph{\textbf{Decomposition Quality and Claim Granularity.}}  
\citet{wanner-etal-2024-closer} introduce \textsc{DecompScore} to quantify decomposition consistency. \citet{chen2023felmbenchmarkingfactualityevaluation} and \citet{yue-etal-2024-retrieval} report failures tied to vague or anaphoric claims. \citet{hu2025decompositiondilemmasdoesclaim} catalog common decomposition errors and tradeoffs. \veriscore\ explicitly avoids unverifiable content like opinions or suggestions \citep{veriscore}.

\paragraph{\textbf{End-to-End and Lightweight Approaches.}}  
\textsc{MiniCheck} avoids decomposition entirely, using a small model to classify sentence-level factuality \citep{tang2024minicheckefficientfactcheckingllms}. \textsc{LLM-OASIS} introduces a synthetic benchmark for scalable factuality evaluation \citep{scirè2025truthmirageendtoendfactuality}, while \textsc{FactCheck-GPT} trains an end-to-end verifier integrating retrieval and judgment \citep{wang-etal-2024-factcheck}. 

\section{Conclusion}
\label{sec:conclusion}
We present \name, a model that unifies claim decomposition and verification into a single model pass. Trained on synthetic data derived from \veriscore, our method processes long-form outputs with consolidated evidence and produces fine-grained factuality judgments. 

Despite the complexity of the task, \name\ closely matches the accuracy of \veriscore\ (Pearson $r = 0.80$), while achieving a 6.6$\times$ speedup in evaluation time. It also outperforms strong few-shot prompting baselines such as GPT-4o, and provides interpretable, claim-level outputs suitable for evaluation and alignment.

We release the model and data to support scalable, efficient factuality assessment and to facilitate further research on long-form factuality and reward modeling.
\section{Limitations}

\paragraph{Latency remains a bottleneck.}  
While \name\ offers substantial speedups over multi-stage pipelines like \veriscore, it still requires full-sequence generation conditioned on long evidence and response contexts. This makes it less suited for real-time or low-latency feedback settings such as RLHF. Although we explored a scoring-head variant (Appendix~\ref{appendix:scoring_head}) to reduce inference cost, it struggled to match the accuracy and reliability of the generative approach.

\paragraph{Reliance on synthetic supervision.}  
\name\ is trained on synthetic labels produced by \veriscore, which, while effective, is not free of noise or bias. Errors in claim decomposition, incomplete evidence, or incorrect verifiability judgments in the teacher outputs may propagate into the learned model, potentially reinforcing systemic weaknesses in downstream predictions.

\paragraph{Retrieval noise and specificity.}  
Our retrieval strategy issues Google search queries for each sentence in the model response, aggregating the results into a shared evidence context. While this aligns with the single-pass structure of \name, it can introduce irrelevant or insufficient evidence, especially for vague or underspecified sentences. More adaptive or agentic retrieval, allowing the model to detect failures and issue improved follow-up queries, may yield higher evidence quality.

\paragraph{Lack of explicit rationales.}  
\name\ produces fine-grained factuality labels for claims but does not provide natural language justifications. Although rationale generation adds computational cost, it could improve interpretability and trustworthiness, especially in sensitive domains. Training with rationales, even if only used during finetuning (e.g., as in Qwen3’s “reasoning mode” \cite{yang2025qwen3technicalreport}), may offer benefits without incurring runtime overhead at inference time. Initial experiments on training a model to reason first using GRPO showed promise.

\paragraph{Noise in evaluation signals.}  
We use GPT-4o-mini to assess label correctness and semantic equivalence, which enables scalable evaluation but introduces potential noise, particularly for nuanced paraphrase detection. This may affect metric estimates such as precision and recall, and obscure more subtle performance differences across models.

\paragraph{Language and configuration limitations.}  
Our experiments focus primarily on English-language outputs. Although some non-English generations were present, we did not conduct language-specific analysis. Additionally, we only minimally modified prompts from \veriscore\ and did not exhaustively tune training configurations. Future work could explore multilingual robustness and systematically investigate the effects of prompt and hyperparameter choices.

\section*{Ethics Statement}

This work involves training and evaluating factuality metrics for long-form language generation. All model outputs analyzed in this study were generated by publicly available language models. The human evaluation was conducted by the first author and involved no sensitive or private user data. No personally identifiable information (PII) was collected, and no crowdworkers or third-party annotators were employed. The synthetic data used to train our model was generated automatically using existing metrics and publicly available datasets. We will release our model and training data upon publication to support reproducibility and future research.
\section*{Acknowledgments}
We extend our special gratitude to Yixiao Song for sharing data and details about \veriscore\ with us. We also extend gratitude to members from the NGRAM lab at UMD for their feedback. This project was partially supported by awards IIS-
2046248, IIS-2312949, and IIS-2202506 from the
National Science Foundation (NSF).

\bibliography{main}

\appendix
\section{Appendix}
\label{sec:appendix}

\subsection{Training \name}
\label{appendix:training}
    We follow a two-stage pipeline to train our \name\ model. 
    \begin{itemize}
        \item In the \textbf{first stage}, we finetune Llama3.1 8B on the combined train split of \veriscoredata\ and \tulupersonas\ data using claim-level evidence.
        \item In the \textbf{second stage}, we further finetune the best  model from stage 1 on the same dataset but with a mixture (60\%:40\%)  of claim-level and sentence-level evidence.
    \end{itemize}  

\paragraph{Motivation for multi-stage training:} In our synthetic data, ground-truth claim verification labels were obtained from \veriscore\ using claim-level evidence. To ensure correctness of labels and model's reliance on evidence provided in the input during generation of output labels, the first stage trains \name\ model using consolidated claim-level evidence.

\paragraph{Drop in accuracy with sentence-level evidence:}As seen in \tableref{tab:dataset_performance}, there was a $\sim$20\% drop in claim accuracy when using sentence-level evidence, which is our true test setting, instead of claim-level evidence. This is possibly due to differences in quality of retrieved search results and length of consolidated evidence depending on the granularity at which evidence was collected. To mitigate this, the model was further trained on a mixture of claim- and sentence-level evidence in the second stage of training. \\

\paragraph{Compute details:}
We finetune our models for 10 epochs on 4 A100 GPUs for about 1-2 day(s). We evaluate the model after every epoch on our validation split and save the best model based on claim accuracy. Test evaluations of the model were performed on a single GH200 GPU node.

\subsection{Automatic Evaluation}
\label{appendix:auto-eval-appendix}
To evaluate the factuality labels produced by \name, we perform automatic evaluation using GPT-4o-mini as an entailment judge. Given an extracted claim and a set of reference claims, the model is prompted to determine whether the claim can be inferred from the reference set. If supported, the model also identifies a minimal subset of reference claims that together entail the extracted claim. 
More concretely, GPT-4o-mini was prompted to judge factual consistency of each predicted claim against gold claims, and vice versa.
Factual precision is computed from predicted claims given the gold claims as the reference set, while factual recall is computed from gold claims given the predicted claims as the reference set.
This two-way setup accounts for granularity mismatches (e.g., one composite claim vs.\ multiple atomic claims), which precludes one-to-one alignment while also ensuring that factual consistency is assessed in a targeted, evidence-grounded manner while minimizing overreliance on superficial textual overlap.

\paragraph{}For scoring purposes, we use the original \veriscore\ verification labels assigned to the retrieved subset of justifying reference claims. If any of the claims in this minimal set are labeled as Unsupported by \veriscore, the extracted claim is marked as Unsupported for evaluation. This setup enables us to scale the evaluation of thousands of (claim, evidence) pairs with approximate but consistent supervision.

\subsection{\namer\ model}
\label{appendix:scoring_head}

To take the cost-efficiency one-step further, we trained a model with a scoring head to directly produce the factuality score given the model response and consolidated sentence-level evidence without going through the whole process of claim decomposition and verification. We envision such a model being used as a reward model in RLHF setups to train models to produce more factual responses. 

\paragraph{Model architecture:} We experimented with starting from the base model as well as from the best model from our multi-stage training pipeline. We explored two different architecture for the scoring head:
\begin{itemize}
    \item Linear output layer on top of the last hidden state of the last token in the input prompt
    \item Linear projection layer on top of a learnable attention layer over the last hidden state of all tokens in the input prompt followed by a linear output layer
\end{itemize}

To avoid overfitting, we added a dropout layer in both of these configurations and ablated by freezing the language model parameters. We also experimented with a multi-task learning setup with losses computed with both the scoring head output and the decomposition and verification outputs from the language modeling head.

\paragraph{Abysmal generalization:} Our best model achieved 75\% score accuracy on the train split, 55\% on the validation split and 45\% on the test split. Due to  poor correlation with ground-truth scores, this avenue was abandoned. We leave it for future work to explore more complex scoring head architectures and training configurations. 

\onecolumn
\begin{center}
\begin{longtable}{|p{0.98\textwidth}|}
\caption{\textbf{Automatic Evaluation:} Prompt used for automatic evaluation with GPT-4o-mini. Given an extracted claim and a list of reference claims, the task is to judge whether the extracted claim can be inferred from the provided list of claims. If that is true, the model also needs extract a minimal subset of claims from the provided list of claims which entails the extracted claim.}
\label{tab:auto-eval-prompt-format} \\
\hline
\endfirsthead

\hline
\endhead

\hline
\textit{\textbf{Automatic Evaluation:} Continued on next page} \\
\hline
\endfoot

\hline
\endlastfoot

You are a helpful assistant tasked with judging the factuality of a statement given some evidence by retrieving a minimal set of facts in the evidence that is necessary and sufficient to justify your judgement of the factuality of the statement.

A fact is justifying your judgement if and only if the fact needs to be true for your judgement to be true.

Below are the definitions of the two categories:

\textbf{Supported}: A fact is supported by the evidence if everything in the fact is supported and nothing is contradicted by the evidence. Evidence may contain other information not fully related to the fact.

\textbf{Unsupported}: A fact is unsupported by the evidence if something in the fact is either contradicted by some information in the evidence or cannot be verified from the evidence.

\textbf{Input format:}

\#\#\# Evidence

1. <fact 1 here>

2. <fact 2 here>

...

n. <fact n here>

\#\#\# Statement

<statement here>

\textbf{Response format:}

\#\#\# Thoughts

<your thoughts here>

\#\#\# Justifying Facts

<justifying fact number 1>, <justifying fact number 2>, ..., <justifying fact number k>

\#\#\# Judgement

Supported / Unsupported

\textbf{Guidelines:}

Only mention fact numbers as a comma separated list under justifying facts and not the entire text of the claims.

Do not include facts that are not strictly required to be true to justify your judgement of the factuality of the statement.

If there are no justifying facts, return None.
\\
\#\#\# Claims

1. Colleen Hoover's \textit{Without Merit} book is a work by Colleen Hoover.

2. Several titles by Colleen Hoover have become \textit{New York Times} bestsellers.

3. Colleen Hoover's \textit{Heart Bones} book is a work by Colleen Hoover.

4. Colleen Hoover's \textit{Slammed} series includes 3 books.

5. Colleen Hoover has published 24 books as of February 2023.

6. Colleen Hoover's books include series.

7. Colleen Hoover's \textit{All Your Perfects} book is a work by Colleen Hoover.

8. Colleen Hoover's \textit{Reminders of Him} book is a work by Colleen Hoover.

9. Colleen Hoover's \textit{Hopeless} series includes 2 books.
\\
10. Colleen Hoover's books have gained immense popularity in recent years.

11. Colleen Hoover is a prolific writer in the New Adult genre.

12. Colleen Hoover's \textit{Layla} book is a work by Colleen Hoover.

13. Colleen Hoover's \textit{It Starts with Us} book is a sequel to \textit{It Ends with Us}.

14. Colleen Hoover's \textit{It Ends with Us} book is a work by Colleen Hoover.

15. Colleen Hoover is a prolific writer in the romance genre.

16. Colleen Hoover's books include standalone novels.

17. Colleen Hoover's books include novellas.

\#\#\# Statement

Colleen Hoover has published a book titled “Hopeless” as part of the “Hopeless” series.

\end{longtable}
\end{center}

\clearpage
\twocolumn

\begin{table*}[t]
\centering
\fontsize{14}{18}\selectfont
\resizebox{\textwidth}{!}{%
\begin{tabular}{@{}p{4.5cm}p{4.5cm}C{1.75cm}C{1.5cm}ccccp{1cm}@{}}
\toprule
{\begin{tabular}[c]{@{}l@{}}\textbf{Data Source,}\\\textbf{Evidence Granularity} \end{tabular}}
  & \textbf{Model} 
  & \textbf{Precision $\uparrow$} 
  & \textbf{Recall $\uparrow$} 
  & \multicolumn{3}{c}{\textbf{Claim Accuracy (\%)}} 
  & \textbf{Correlation b/w factuality scores $\uparrow$} \\ 
\cmidrule(lr){5-7}
& & & & \textbf{Correct $\uparrow$} & \textbf{Incorrect $\downarrow$} & \textbf{Missing $\downarrow$} & \\
\midrule

\multirow{3}{*}{\begin{tabular}[c]{@{}l@{}}\veriscore\ \&\\ Tulu3 Personas,\\claim‐level\end{tabular}}
  & GPT-4o few-shot ICL     & 0.10    & 0.04    & 3.54    & \textbf{0.82}    & 95.64    & 0.28    \\
& \name-\textit{Stage-1}     & \textbf{0.06}    & \textbf{0.29}    & \textbf{24.64}    & 4.53    & \textbf{70.84}    & 0.85    \\
\cmidrule(lr){2-8}
& \name  & 0.05 & 0.28 & 23.69 & 4.35 & 71.96 & \textbf{0.87} \\
\midrule

\multirow{3}{*}{\begin{tabular}[c]
{@{}l@{}}Tulu3 Personas,\\sentence‐level\end{tabular}}
& GPT-4o few-shot ICL     & \textbf{0.15}    & 0.03    & 2.61    & \textbf{0.57}    & 96.82    & 0.33    \\
& \name-\textit{Stage-1}     & 0.06    & 0.19    & 15.33    & 3.43    & 81.24    & 0.75    \\
\cmidrule(lr){2-8}
& \name  & 0.05 & \textbf{0.23} & \textbf{18.62} & 4.18 & \textbf{77.26} & \textbf{0.80} \\
\bottomrule
\end{tabular}%
}
\caption{\textbf{Exact match metrics for different test splits.} \name-\textit{Stage-1} is the best model from Stage 1 training with claim-level evidence. \textit{Evidence Granularity} indicates whether provided evidence was collected using claims or sentences in the model responses as search queries. \textit{Precision} measures how many claims produced by \name\ is also produced by \veriscore. \textit{Recall} measures how many claims produced by \veriscore is also produced by \name. Under \textit{Claim accuracy}, \textit{Correct} measures the accuracy of the verification label produced by \name\ with label produced by \veriscore\ as the ground truth. \textit{Incorrect (Label)} measures how many claims produced by both \name\ and \veriscore\ but with differing labels. \textit{Missing (Claim)} measures how many claims produced by \veriscore\ but not by \name. \textit{ Correlation b/w factuality scores} is the Pearson's correlation between factuality scores computed by  \name\ and \veriscore\ for model responses.}
\label{tab:dataset_performance}
\end{table*}

\begin{table*}[t]
\centering
\fontsize{14}{18}\selectfont
\resizebox{\textwidth}{!}{%
\begin{tabular}{@{}p{4.5cm}p{4.5cm}C{1.75cm}C{1.5cm}ccccp{1cm}@{}}
\toprule
{\begin{tabular}[c]{@{}l@{}}\textbf{Data Source,}\\\textbf{Evidence Granularity} \end{tabular}}
  & \textbf{Model} 
  & \textbf{Precision $\uparrow$} 
  & \textbf{Recall $\uparrow$} 
  & \multicolumn{3}{c}{\textbf{Claim Accuracy (\%)}} 
  & \textbf{Correlation b/w factuality scores $\uparrow$} \\ 
\cmidrule(lr){5-7}
& & & & \textbf{Correct $\uparrow$} & \textbf{Incorrect $\downarrow$} & \textbf{Missing $\downarrow$} & \\
\midrule

\multirow{3}{*}{\begin{tabular}[c]{@{}l@{}}\veriscoredata\ \&\\ \tulupersonas,\\claim‐level\end{tabular}}
  & GPT-4o few-shot ICL     & 0.30    & 0.31    & 18.1    & \textbf{7.2}    & 73.2    & 0.28    \\
& \name-\textit{Stage-1}     & \textbf{0.88}    & 0.87    & 63.0    & 21.3    & 15.1    & 0.85    \\
\cmidrule(lr){2-8}
& \name  & 0.87 & \textbf{0.90} & \textbf{71.6} & 15.5 & \textbf{12.3} & \textbf{0.87} \\
\midrule

\multirow{3}{*}{\begin{tabular}[c]
{@{}l@{}}\tulupersonas,\\sentence‐level\end{tabular}}
& GPT-4o few-shot ICL     & 0.38    & 0.34    & 19.3    & \textbf{8.2}    & 71.2    & 0.33    \\
& \name-\textit{Stage-1}     & 0.72    & 0.69    & 50.8    & 20.8    & 27.7    & 0.75    \\
\cmidrule(lr){2-8}
& \name  & \textbf{0.83} & \textbf{0.86} & \textbf{66} & 17.6 & \textbf{15} & \textbf{0.80} \\
\bottomrule
\end{tabular}%
}
\caption{\textbf{Automatic performance metrics across test splits.} \name-\textit{Stage-1} refers to the best model trained with claim-level evidence. \textit{Evidence Granularity} denotes whether evidence was retrieved using full sentences or claims. \textit{Precision} and \textit{Recall} measure overlap between claims identified by \name\ and \veriscore. Under \textit{Claim Accuracy}, \textit{Correct}, \textit{Incorrect (Label)}, and \textit{Missing (Claim)} capture agreement, label mismatches, and omissions relative to \veriscore. A small fraction of instances were marked \textit{Erroneous} (not shown) due to ambiguities in automatic evaluation (see \ref{appendix:auto-eval-appendix}). \textit{Correlation} reports Pearson $r$ between factuality scores (see  Eq. \ref{eq:factuality_score}) assigned by \name\ and \veriscore. For all Pearson's correlation scores reported in this paper, p<0.001, unless stated otherwise.}

\label{appendix:dataset_performance}
\end{table*}

\label{appendix:prompt}
\onecolumn
\begin{center}
\begin{longtable}{|p{0.98\textwidth}|}
\caption{\textbf{\name\ prompt format:} Sample input and output prompt}
\label{tab:prompt-format} \\
\hline
\endfirsthead

\hline
\endhead

\hline
\textbf{\name\ prompt format:} \textit{Continued on next page} \\
\hline
\endfoot

\hline
\textbf{\name\ prompt format} \\
\hline
\endlastfoot
{\itshape
You are trying to verify how factual a response is by extracting fine-grained, verifiable claims. \newline
Each claim must describe one single event or one single state (for example, “Nvidia was founded in 1993 in Sunnyvale, California, U.S.”) in one sentence with at most one embedded clause. \newline
Each fact should be understandable on its own and require no additional context. \newline
This means that all entities must be referred to by name but not by pronoun. \newline
Use the name of entities rather than definite noun phrases (e.g., “the teacher”) whenever possible. \newline
If a definite noun phrase is used, be sure to add modifiers (e.g., an embedded clause or a prepositional phrase). \newline
Each fact must be situated within relevant temporal and location details whenever needed. \newline

All necessary specific details-including entities, dates, and locations—must be explicitly named, and \textit{verify} here means that every detail of a claim is directly confirmed by the provided evidence. \newline
The verification process involves cross-checking each detail against the evidence; a detail is considered verified if it is clearly confirmed by the evidence. \newline

Avoid extracting stories, personal experiences, hypotheticals (e.g., those using “would be” or the subjunctive mood), subjective opinions, suggestions, advice, instructions, or similarly non-factual content; however, biographical, historical, scientific, and similar texts are acceptable. \newline
Also, ignore any listed references. \newline

For each extracted claim, classify it as follows: \newline

\textbf{Supported}: Every detail of the claim (including entities, dates, and locations) is directly confirmed by the provided evidence with no contradictions. \newline
\textbf{Unsupported}: One or more details of the claim are either missing from or contradicted by the provided evidence, even though the claim remains verifiable using external sources. \newline

You do not need to justify what you extract. \newline

Output format: \newline
<fact 1>: <your judgment of fact 1> \newline
<fact 2>: <your judgment of fact 2> \newline
… \newline
<fact n>: <your judgment of fact n> \newline

If no verifiable claim can be extracted, simply output "No verifiable claim." \newline}
{\itshape\textbf{\#\#\# Response:}}
\newline
\textbf{Historical Context and Significance:}In the summer of 1215, a pivotal moment in the course of Western political and legal history unfolded in England. Amidst mounting discontent between King John and his barons, a document was drafted in Runnymede, a meadow along the River Thames. This document, known as the Magna Carta, was the first significant attempt to limit the power of a monarch by law, and it marked a crucial step towards modern representative democracy.\newline

\textbf{The Heart of the Matter: The Magna Carta's Provisions:}The Magna Carta, officially known as the "Great Charter", was a compact between King John and his barons, sealing a series of agreements that aimed to secure political rights and liberties. Some of the most influential clauses protected church rights, granted trial by jury, limited the amount of unjust taxes, and established the principle that no free man could be imprisoned or exiled without a fair trial. These provisions laid the foundation for the rule of law, constitutional government, and individual rights that are widely upheld today.
\\
\textbf{Lasting Impact and Legacy:}The Magna Carta has left an indelible mark on the annals of history and continues to serve as a symbol of liberty and the rule of law in the modern world. Although it has undergone numerous reissues and modifications throughout the centuries, the original 1215 charter remains a significant historical artifact. It has been celebrated as a precursor to the Bill of Rights in the United States, the Universal Declaration of Human Rights, and many other international charters and constitutions. More than eight centuries after its inception, the Magna Carta's enduring legacy continues to inspire efforts towards justice, freedom, and governance worldwide.\newline

{\itshape\textbf{\#\#\# Evidence}} \newline
The Magna Carta is the charter of English liberties granted by King John on June 15, 1215, under threat of civil war.\newline
The Magna Carta (or Great Charter) was written in Latin and was effectively the first written constitution in European history.\newline
This significant moment, the first time a ruling monarch had been forcibly persuaded to renounce a great deal of his authority, took place at Runnymede, a ...\newline
It promised the protection of church rights, protection for the barons from illegal imprisonment, access to swift and impartial justice, and limitations on ...\newline
Eight hundred years ago today, King John of England sealed the Magna Carta, a groundbreaking legal document that served as the foundation for ...\newline
The Magna Carta, officially granted by King John of England on 15 June 1215, stands as one of the most influential and pivotal documents in ...\newline
On June 15, 1215, in a field at Runnymede, King John affixed his seal to Magna Carta. Confronted by 40 rebellious barons, he consented to their ...\newline
The Department of Defense undertook preparation ofa history ofthe 11. September 2001 attack on the Pentagon at the suggestion of Brig. Gen. John S. Brown, ...\newline
hours This course examines the causes and consequences of globalization. Issues are examined from a changing historical context of economy, politics, and ...\newline
\newline
... along the south bank of the Thames River, in a meadow called Runnymede. King John affixed his seal to the Magna Carta on June 15, 1215. The document then ...\newline
On June 15, 1215, John met the barons at Runnymede on the Thames and set his seal to the Articles of the Barons, which after minor revision was ...\newline
The result was the Magna Carta, which was signed on June 15, 1215, at Runnymede, a meadow by the River Thames. Key Provisions of the Magna Carta. While the ...\newline
They accrued at Runnymede, a meadow alongside the banks of the River Thames, on a fateful day in June. They demanded that King John renowned their rights and ...\newline
Magna Carta originated in 1215 as a peace treaty between King John and a group of rebellious barons · The original document was written in Latin ...\newline
A deal was brokered between King John and the barons who agreed to meet on 10th June 1215 at Runnymede, a large flat expanse of meadow near Egham, ideally ...\newline
In May the barons took London, and withdrew their homage and fealty. In the middle of June, 1215, on a meadow, Runnymede, along the River Thames the rebellious ...\newline
McKechnie's introduction to Magna Carta: A Commentary on the Great Charter of King John, with an Historical Introduction, by William Sharp McKechnie.\newline
The story culminates at Runnymede, a meadow on the banks of the River Thames. ... by King John in 1215, was a document of unprecedented ... its origins in the ...\newline
\newline
The Magna Carta is the charter of English liberties granted by King John on June 15, 1215, under threat of civil war.\newline
\\
Magna Carta was issued in June 1215 and was the first document to put into writing the principle that the king and his government was not above the law.\newline
In the early 17th century, Magna Carta became increasingly important as a political document in arguments over the authority of the English monarchy.\newline
June 15, 2015 marks the 800th anniversary of the Magna Carta, a pioneering legal document that served as the foundation for American democracy and individual ...\newline
The document, often referred to simply as Magna Carta, did indeed innumerate several various rights and liberties, as well as limitations on the King's powers ...\newline
Magna Carta is significant because it is a statement of law that applied to the kings as well as to his subjects.\newline
The Magna Carta was the first step of the English Monarchy giving up some power and allowing the Judicial Branches have more power. In theory it ...\newline
The Magna Carta established that the king and all individuals were subject to the law, codified the principles of due process and trial by jury.\newline
Magna Carta marked an important step, in the process by which England became a nation; but that step was neither the first nor yet the final one.2. In ...\newline
\newline
The Magna Carta, Latin for “Great Charter,” is one of the most famous documents in history. It was originally issued in 1215 during the reign of ...\newline
A law made by the king in one national assembly might be repealed by the king in another; whereas the Great Charter was intended by the barons to be ...\newline
MCMXIV. "vo. Page 7. MAGNA CARTA. A COMMENTARY ON THE GREAT. CHARTER OF KING JOHN. WITH AN. HISTORICAL INTRODUCTION. BY. WILLIAM SHARP McKECHNIE ... Great Charter ...\newline
Magna Carta: A Commentary on the Great Charter of King John, with an Historical Introduction, by William Sharp McKechnie (Glasgow: Maclehose, 1914). Copyright.\newline
Full text of "Magna carta : a commentary on the Great Charter of King John with an historical introduction" ... Great Charter was intended by the barons to be ...\newline
a Commentary on the Great Charter of King John (2nd edn., Glasgow, 1914). ... have all their rights and liberties according to the great charter of England'.\newline
During his final year of public service, Coke forced a reconsideration of the Great Charter's overall meaning and promise. Would the liberties of the.\newline
But minding and mining the Magna Carta became a fun task, not just to learn about the Great Charter, but to glean from his notes or his citations the major ...\newline
called "the statute r :lied the Great Charter of the Liberties of England." The simple fact that the language of Maps Carta bestowed its benefits on "free ...\newline
Indeed, it was named the 'Great Charter', several years after the initial 1215 settlement, not in recognition of its importance, but because it ...\newline
\newline
It promised the protection of church rights, protection from illegal imprisonment, access to swift justice, and, most importantly, limitations on taxation and ...\newline
Of enduring importance to people appealing to the charter over the last 800 years are the famous clauses 39 and 40: “No free man shall be seized, imprisoned, ...\newline
The two most famous clauses; establishing the right of all to be judged by their equals, and outlawing imprisonment of free men without a trial, were clauses 39 ...\newline
Among these are the principle of no taxation without representation and the right to a fair trial under law.\newline
Clause 12 prevented kings from imposing taxes 'without common counsel'. The principle – that taxation must be by consent – became fixed in English politics.\newline
\\
The English church shall be free and shall have its rights intact and its liberties uninfringed upon. And thus we will that it be observed.\newline
... No free man shall be seized or imprisoned, or stripped of his rights or possessions, or outlawed or exiled, or deprived of his standing in any other way, nor.\newline
+ (39) No free man shall be seized or imprisoned, or stripped of his rights or possessions, or outlawed or exiled, or deprived of his standing in any way, nor ...\newline
These guaranteed the Church the freedom to handle its business without interference from the throne; that no free man could be imprisoned or outlawed except by ...\newline
One of the most important, and often quoted, provisions stated that "no freeman shall be seized, imprisoned, dispossessed, outlawed, or exiled, or in any way ...\newline
\newline
The Federalist \# 78 states further that, if any law passed by Congress conflicts with the Constitution, "the Constitution ought to be preferred to the statute, ...\newline
The Bill of Rights is a founding documents written by James Madison. It makes up the first ten amendments to the Constitution including freedom of speech ...\newline
Article III of the Constitution establishes the federal judiciary. Article III, Section I states that "The judicial Power of the United States, shall be vested ...\newline
The Bill of Rights became a document that defends not only majorities of the people against an overreaching federal government but also minorities against ...\newline
"[A] bill of rights is what the people are entitled to against every government on earth, general or particular, and what no just government ...\newline
The higher law, reciprocity and mutuality of obligations, written charters of rights, the right to be consulted on policy and to grant or refuse one's consent, ...\newline
The. Supreme Court upheld the law, ruling that Congress has the power to enact laws that directly affect the acts of individuals, thereby making ...\newline
The Universal Declaration of Human Rights is generally agreed to be the foundation of international human rights law. Adopted in 1948, the UDHR has inspired ...\newline
Article III of the Constitution establishes and empowers the judicial branch of the national government.\newline
Today, the rule of law is often linked to efforts to promote protection of human rights worldwide.\newline
\newline
Magna Carta was the result of the Angevin king's disastrous foreign policy and overzealous financial administration.\newline
Magna Carta's clauses provided the basis for important principles in English law developed in the fourteenth through to the seventeenth century.\newline
Magna Carta was written by a group of 13th-century barons to protect their rights and property against a tyrannical king.\newline
Though the king never meant to keep his promises, Magna Carta survived. Down through the centuries, it has been a symbol of opposition to arbitrary government.\newline
The Magna Carta is the charter of English liberties granted by King John on June 15, 1215, under threat of civil war.\newline
The Magna Carta (or Great Charter) was written in Latin and was effectively the first written constitution in European history.\newline
The Magna Carta, the medieval English historic legal document that is seen as the origin of many modern-day legal rights and constitutional principles.\newline
It is a story that runs 800 years forward and is still unfolding. It is the story of our rule of law tradition and of how our American system of government is ...\newline
exploration of how historical events have left an indelible mark on present-day national political systems, employing a comparative analysis of the British ...\newline
\\
None of the original 1215 Magna Carta is currently in force since it has been repealed; however, four clauses of the original charter are enshrined in the 1297 ...\newline
On the back of this charter issued in the name of King Æthelbald of Mercia in 736, it is still possible to see the impressions from where the document was ...\newline
The king and the rebel barons negotiated a peace settlement in June 1215. The king agreed to accept the terms of Magna Carta, which is dated 15 June 1215.\newline
David Carpenter considers the huge significance of the 13th-century document that asserted a fundamental principle – the rule of law.\newline
That any copies of the 1215 Magna Carta survive is even more remarkable given the number of times it was reissued—most copyholders would ...\newline
Magna Carta, charter of English liberties granted by King John on June 15, 1215, under threat of civil war and reissued, with alterations, in 1216, 1217, and ...\newline
Such an evaluation will not prove that everything commonly said in praise of the 1215 Charter is true. Some of it is myth. But it is not all myth. Many of the ...\newline
Representing a would-be peace treaty between the king and rebellious nobles, the 1215 Charter did not survive its year of issue. Pope Innocent ...\newline
The Magna Carta is a document written in 1215 that limited the power of the English king. It remains important today as a symbol of protest against ...\newline
This version is arguably just as significant as the 1215 charter in that it remained materially unchanged through various reissues and confirmations until in ...\newline
\newline
The UDHR is widely recognized as having inspired, and paved the way for, the adoption of more than seventy human rights treaties, applied today on a permanent ...\newline
It directly inspired the development of international human rights law, and was the first step in the formulation of the International Bill of Human Rights, ...\newline
The Universal Declaration of Human Rights, which was adopted by the UN General Assembly on 10 December 1948, was the result of the experience of the Second ...\newline
These three documents, known collectively as the Charters of Freedom, have secured the rights of the American people for more than two and a quarter centuries.\newline
The Magna Carta established the rule of law and the idea that all citizens, including those in power, should be fairly and equally ruled by the law.\newline
This exhibit celebrates the leadership of Eleanor Roosevelt in writing the Universal Declaration of Human Rights as we mark the 70 th anniversary of its ...\newline
The Universal Declaration of Human Rights has been the centrepiece of the modern international law of human rights for more than sixty years. If anything ...\newline
Everyone is entitled to all the rights and freedoms set forth in this Declaration, without distinction of any kind, such as race, colour, sex, language, ...\newline
Known today as the Cyrus Cylinder, this ancient record has now been recognized as the world's first charter of human rights. It is translated into all six ...\newline
The Universal Declaration of Human Rights describes human rights of all people around the world. D. The Bill of Rights to the U.S. Constitution ...\newline
\newline
In 1215, when King John confirmed Magna Carta with his seal, he was acknowledging the now firmly embedded concept that no man--not even the king ...\newline
As we celebrate 800 years of Magna Carta, it's timely to reflect on the bloodshed and tyranny behind its creation and the reasons for its ...\newline
June 15, 2015 marks the 800th anniversary of the Magna Carta, a pioneering legal document that served as the foundation for American democracy and individual ...\newline
\\
The story of Magna Carta began in 1215, but it continues eight centuries forward and is still unfolding. It is the story of modern constitutional government and ...\newline
Magna Carta established a number of very important political principles that framed and informed our discussions about human rights in the centuries that ...\newline
The Magna Carta is the charter of English liberties granted by King John on June 15, 1215, under threat of civil war.\newline
A Magna Carta moment. After eight centuries the revered document of liberty still grips the political imagination, says David Hayes in London.\newline
Magna Carta was not made law; it was more of a working document setting out how the country would be run.\newline

{\itshape\textbf{\#\#\# Facts}} \newline
The Magna Carta serves as a symbol of the rule of law.: Supported\newline
The Magna Carta is celebrated as a precursor to the Bill of Rights in the United States.: Supported\newline
The Magna Carta is officially known as the "Great Charter".: Unsupported\newline
The Magna Carta's legacy continues to inspire efforts towards freedom worldwide.: Unsupported\newline
The Magna Carta's legacy continues to inspire efforts towards justice worldwide.: Supported\newline
The provisions of the Magna Carta laid the foundation for constitutional government.: Supported\newline
Individual rights are widely upheld today.: Unsupported\newline
The Magna Carta has influenced many international charters and constitutions.: Supported\newline
The Magna Carta protected church rights.: Supported\newline
The Magna Carta established the principle that no free man could be imprisoned or exiled without a fair trial.: Supported\newline
The provisions of the Magna Carta laid the foundation for individual rights.: Supported\newline
The Magna Carta was the first significant attempt to limit the power of a monarch by law.: Supported\newline
The Magna Carta has left an indelible mark on the annals of history.: Supported\newline
The Magna Carta's legacy continues to inspire efforts towards governance worldwide.: Supported\newline
The provisions of the Magna Carta laid the foundation for the rule of law.: Supported\newline
The Magna Carta is celebrated as a precursor to the Universal Declaration of Human Rights.: Supported\newline
The drafting of the Magna Carta was amidst mounting discontent between King John and his barons.: Supported\newline
The Magna Carta was a compact between King John and his barons.: Supported\newline
Constitutional government is widely upheld today.: Unsupported\newline
The Magna Carta is considered a pivotal moment in Western political and legal history.: Supported\newline
The Magna Carta has influenced the Universal Declaration of Human Rights.: Supported\newline
The Magna Carta took place in England.: Supported\newline
The Magna Carta aimed to secure political rights and liberties.: Supported\newline
The Magna Carta granted trial by jury.: Supported\newline
The Magna Carta was incepted more than eight centuries ago.: Unsupported\newline
The drafting of the Magna Carta took place in the summer of 1215.: Unsupported\newline
The Magna Carta was drafted in the summer of 1215.: Unsupported\newline
The Magna Carta serves as a symbol of liberty.: Supported\newline
The Magna Carta has influenced the Bill of Rights in the United States.: Supported\newline
The rule of law is widely upheld today.: Unsupported\newline
The original 1215 charter of the Magna Carta remains a significant historical artifact.: Unsupported\newline
The Magna Carta was drafted in Runnymede, a meadow along the River Thames.: Supported\newline
\\
The Magna Carta marked a crucial step towards modern representative democracy.: Supported\newline
The Magna Carta limited the amount of unjust taxes.: Unsupported\newline
The Magna Carta is celebrated as a precursor to many international charters and constitutions.: Supported\newline
The original 1215 charter of the Magna Carta has undergone numerous reissues and modifications throughout the centuries.: Supported\newline
\end{longtable}
\end{center}
\clearpage
\twocolumn

\onecolumn
\label{appendix:tulu3-sieve}
\begin{center}
\begin{longtable}{|p{0.98\textwidth}|}
\caption{\textbf{Automatic Filtering:} Prompt used for automatic filtering of Tulu3-personas prompts with GPT-4o-mini.}
\label{tab:auto-eval-prompt-format} \\
\hline
\endfirsthead

\hline
\endhead

\hline
\textit{\textbf{Automatic Filtering:} Continued on next page} \\
\hline
\endfoot

\hline
\endlastfoot

You are a helpful assistant. Given a prompt, you are to judge whether a response to the prompt would contain fine-grained factually verifiable claims. Each of these fine-grained facts should be verifiable against reliable external world knowledge (e.g., via Wikipedia). 

Any story, personal experiences, hypotheticals (e.g., "would be" or subjunctive), subjective statements (e.g., opinions), suggestions, advice, instructions, and other such content is considered factually verifiable. 
Biographical, historical, scientific, and other such texts are not personal experiences or stories.
\newline
\newline
Here are some examples of responses and some fine-grained factually verifiable claims present in those responses:\newline
\newline
{\itshape
Text:\newline
The sweet potato or sweetpotato (Ipomoea batatas) is a dicotyledonous plant that belongs to the bindweed or morning glory family, Convolvulaceae. Its large, starchy, sweet-tasting tuberous roots are used as a root vegetable. The young shoots and leaves are sometimes eaten as greens.\newline
Sentence to be focused on:\newline
Its large, starchy, sweet-tasting tuberous roots are used as a root vegetable.\newline
Facts:
\begin{itemize}
    \item Sweet potatoes' roots are large.
    \item Sweet potatoes' roots are starchy.
    \item Sweet potatoes' roots are sweet-tasting.
    \item Sweet potatoes' roots are tuberous.
    \item Sweet potatoes' roots are used as a root vegetable.
\end{itemize}
\vspace{0.25cm}
Text:\newline
After the success of the David in 1504, Michelangelo’s work consisted almost entirely of vast projects. He was attracted to these ambitious tasks while at the same time rejecting the use of assistants, so that most of these projects were impractical and remained unfinished.\newline
Sentence to be focused on:\newline
After the success of the David in 1504, Michelangelo’s work consisted almost entirely of vast projects.\newline
Facts:
\begin{itemize}
    \item Michelangelo achieved the success of the David in 1504.
    \item After 1504, Michelangelo’s work consisted almost entirely of vast projects.
\end{itemize}
\vspace{0.25cm}
Text:\newline
 After the success of the David in 1504, Michelangelo’s work consisted almost entirely of vast projects. He was attracted to these ambitious tasks while at the same time rejecting the use of assistants, so that most of these projects were impractical and remained unfinished. In 1504 he agreed to paint a huge fresco for the Sala del Gran Consiglio of the Florence city hall to form a pair with another just begun by Leonardo da Vinci. Both murals recorded military victories by the city (Michelangelo’s was the Battle of Cascina), but each also gave testimony to the special skills of the city’s much vaunted artists.\newline
Sentence to be focused on:\newline}
\\
{\itshape
In 1504 he agreed to paint a huge fresco for the Sala del Gran Consiglio of the Florence city hall to form a pair with another just begun by Leonardo da Vinci.\newline
Facts:
\begin{itemize}
    \item In 1504, Michelangelo agreed to paint a huge fresco for the Sala del Gran Consiglio of the Florence city hall.
    \item Around 1504, Leonardo da Vinci just began with a mural for the Florence city hall.
\end{itemize}
\vspace{0.25cm}
Text:\newline
After the success of the David in 1504, Michelangelo’s work consisted almost entirely of vast projects. He was attracted to these ambitious tasks while at the same time rejecting the use of assistants, so that most of these projects were impractical and remained unfinished. In 1504 he agreed to paint a huge fresco for the Sala del Gran Consiglio of the Florence city hall to form a pair with another just begun by Leonardo da Vinci. Both murals recorded military victories by the city (Michelangelo’s was the Battle of Cascina), but each also gave testimony to the special skills of the city’s much vaunted artists. Leonardo’s design shows galloping horses, Michelangelo’s active nudes—soldiers stop swimming and climb out of a river to answer an alarm.\newline
Sentence to be focused on:\newline
Both murals recorded military victories by the city (Michelangelo’s was the Battle of Cascina), but each also gave testimony to the special skills of the city’s much vaunted artists.\newline
Facts:
\begin{itemize}
    \item Michelangelo’s murals for the Florence city hall recorded military victories by the city.
    \item Leonardo da Vinci’s murals for the Florence city hall recorded military victories by the city.
    \item Michelangelo’s mural for the Florence city hall was the Battle of Cascina.
\end{itemize}
\vspace{0.25cm}
{\itshape 
Text:\newline
I (27f) and my fiance "Leo" (27m) decided to let my FSIL "Maya" (32f) stay at our house because she needed space from her husband due to some relationship struggles they're having. Leo and I had gotten wedding cake samples from an expensive bakery specializing in wedding cakes. We planned to test them along with Maya after we finished up some other wedding plans yesterday. However, when I came home from work to see Leo yelling at Maya, the box the samples came in wide open on the living room table, and Maya arguing with him. I asked what was happening, and Leo angrily told me that while we were both at work, Maya had some friends over and they ended up eating almost all of our cake samples.\newline
Sentence to be focused on:\newline
However, when I came home from work to see Leo yelling at Maya, the box the samples came in wide open on the living room table, and Maya arguing with him.\newline
Facts:\newline
No verifiable claim.
\newline
\newline
Text:\newline
I was a catholic school kid, educated by nuns and somehow on a spring day in 1972, I was called down to the principal’s office by Sister Mary Roberts, who informed me that I had gained admission to Stuyvesant High School. I was excited to be freshman in one of New York City’s elite public schools but soon came to realize that my catholic school education did not provide the groundwork for abstract concepts like science and algebra. My parochial education in Science at St. Joseph’s was essentially “God made it, what else do you need to know?”\newline}}
\\
{\itshape Sentence to be focused on:\newline
I was excited to be freshman in one of New York City’s elite public schools but soon came to realize that my catholic school education did not provide the groundwork for abstract concepts like science and algebra.\newline
Facts:
\begin{itemize}
    \item Stuyvesant High School is in New York City.
    \item Stuyvesant High School is an elite high school.
    \item Stuyvesant High School is a public school.
    \item In 1972, St. Joseph's catholic school education did not provide the groundwork for abstract concepts like science and algebra.
\end{itemize}
\vspace{0.25cm}
Text: \newline
Ãcariya Mun related the story of a dhutanga monk (ascetic monk) who inadvertently went to stay in a forest located next to a charnel ground. He arrived on foot at a certain village late one afternoon and, being unfamiliar with the area, asked the villagers where he could find a wooded area suitable for meditation. They pointed to a tract of forest, claiming it was suitable, but neglected to tell him that it was situated right on the edge of a charnel ground. They then guided him to the forest, where he passed the first night peacefully. On the following day he saw the villagers pass by carrying a corpse, which they soon cremated only a short distance from where he was staying.\newline
Sentence to be focused on:\newline
They then guided him to the forest, where he passed the first night peacefully.\newline
Facts:\newline
No verifiable claim.}
\newline
\newline
Input format:\newline
\#\#\# Prompt\newline
<prompt here>\newline
\newline
Output format:\newline
\#\#\# Thoughts\newline
<your thoughts here>\newline
\newline
\#\#\# Judgement\newline
Yes/No
\end{longtable}
\end{center}

\clearpage
\twocolumn

\end{document}